\documentclass[english,,compatible,ruled,vlined]{spie}
\usepackage[utf8]{inputenc}
\usepackage{color}
\usepackage{calc}
\usepackage{algorithm2e}
\usepackage{amsmath}
\usepackage{graphicx}

\makeatletter

\providecommand{\tabularnewline}{\\}

\usepackage{babel}

\usepackage{multicol}
\usepackage{subfigure}

\usepackage{babel}

\usepackage{babel}

\makeatother

\usepackage{babel}
\begin{document}

\author{Jincheng Zhang\supit{a}, Andrew R. Willis\supit{a} and Jamie
Godwin\supit{b} \skiplinehalf \supit{a}University of North
Carolina at Charlotte, 9201 University City Blvd., Charlotte, NC~~28223
\\
 \supit{b}Air Force Research Laboratory, Munitions Directorate,
Eglin AFB, FL, 32542 \\
 }

\authorinfo{Further author information: (Send correspondence to A. Willis)\\
 A. Willis: E-mail: arwillis@uncc.edu, Telephone: 1 704 687 8420}

\title{Compute-Bound and Low-Bandwidth Distributed 3D Graph-SLAM}
\maketitle
\begin{abstract}
This article describes a new approach for distributed 3D SLAM map
building. The key contribution of this article is the creation of
a distributed graph-SLAM map-building architecture responsive to bandwidth
and computational needs of the robotic platform. Responsiveness is
afforded by integration of a 3D point cloud to plane cloud compression
algorithm that approximates dense 3D point cloud using local planar
patches. Compute bound platforms may restrict the computational duration
of the compression algorithm and low–bandwidth platforms can restrict
the size of the compression result. The backbone of the approach is
an ultra-fast adaptive 3D compression algorithm that transforms swaths
of 3D planar surface data into planar patches attributed with image
textures. Our approach uses DVO SLAM, a leading algorithm for 3D
mapping, and extends it by computationally isolating map integration
tasks from local Guidance, Navigation and Control tasks and includes an addition
of a network protocol to share the compressed plane clouds. The joint
effect of these contributions allows agents with 3D sensing capabilities
to calculate and communicate compressed map information commensurate
with their on-board computational resources and communication channel
capacities. This opens SLAM mapping to new categories of robotic platforms
that may have computational and memory limits that prohibit other
SLAM solutions. 
\end{abstract}


\section{Introduction}

The SLAM problem was introduced nearly 30 years ago \cite{Smith:1986:RES:33838.33842}
and continues to be a fundamental challenge in robotics today. Solutions
to the SLAM problem require algorithms to solve three challenging
problems: (1) how to efficiently recognize salient geometry, (2) how
to associate and integrate map data, and (3) how to track the position
and orientation of the agents in the map. Use of SLAM to generate high-quality
2D or 3D maps is a classical subject and recent work focuses on real-time applications of this algorithm in distributed environments to generate and  updates large-scale maps. Although there are many state-of-the-art RGBD SLAM solutions \cite{newcombe2011kinectfusion,salas2013slam++,Dai2016,whelan2016elasticfusion}, there remain two key shortcomings: the high computational cost (often requiring GPU acceleration) and bandwidth requirements. The requirements prohibit SLAM from being used in robots that having limited computational or memory resources, e.g., light-duty UAVs and swarm-style robots. The computational bottleneck comes from three sources: (1) tracking the robot pose via the RGBD camera data, (2) updating local and global graph-SLAM map with new RGBD map information and (3) detecting loop closures and updating the SLAM map to include detected closures\cite{Aulinas:2008:SPS:1566899.1566949}. Our goal is to address the SLAM computational challenge by leveraging and improving upon a leading method for state of the art 3D SLAM algorithms\cite{Kerl2013}.

As one of the state-of-the-art dense visual SLAM methods for RGBD cameras, DVO SLAM\cite{Kerl2013} represents a leading approach that outperforms many current competing methods\cite{comport2007accurate,steinbrucker2011real}. DVO SLAM incorporate the Direct Visual Odometry (DVO) algorithm to solve the RGBD camera tracking problem. Visual odometry seeks to compute the changes in RGBD camera 3D pose from sensed RGBD images from the camera. The DVO algorithm poses this problem as a search for the pose that minimizes the point cloud photometric (intensity) and depth differences. In conjunction with robust outlier weighting, DVO continues to set the performance benchmark for RGBD camera tracking. DVO SLAM constructs maps as a connected graph of keyframes where  optimizes this graph using the g\textsuperscript{2}o
framework \cite{5979949}. While DVO SLAM performs well for building
accurate maps in near real-time, it is a stand-alone application to be run on single CPU. Hence, DVO SLAM cannot be used for distributed or collaborative SLAM applications. Further,
DVO SLAM requires large computational and memory resources to process RGBD data in real-time.

In this paper, we extend DVO SLAM\cite{Kerl2013} to distributed scenarios
and simultaneously address the computational burden and bandwidth challenge of using RGBD data. Our approach is to separate real-time essential tasks which we refer to as the SLAM \emph{frontend} from non-essential tasks, e.g., SLAM optimization and map building, which we refer to as the SLAM \emph{backend}. We then minimize communication requirements between the frontend and backend by sharing compressed versions of sensed RGBD data. To summarize, the main contributions
of this paper are: 
\begin{itemize}
\item We de-couple the frontend and backend of graph-SLAM systems thus our
approach has unprecedented flexibility by allowing agents to create on-board maps (frontend and backend) or contribute to other maps (only frontend) or only integrate maps (only backend) using over-network SLAM.
\item We propose a new compression algorithm that transforms 3D RGBD point cloud data into
planar patches that compress the geometric data to reduce computational and memory resource requirements for DVO SLAM map building. 
\item We integrate parameters to our compression algorithm so it can adapt to satisfy dynamic computational and/or bandwidth constraints as specified by the host computer. This brings makes it possible to run this state-of-the-art SLAM approach on hosts with limited resources. 
\item We incorporate plane cloud surface representations as a new elements for keyframe-based mapping having many of the benefits of RGBD keyframes with smaller computational cost and memory usage.
\end{itemize}
These contributions significantly advance the state
of the art for RGBD 3D SLAM. By taking the advanced algorithms that
are used by many researchers and advancing it to run
in distributed applications with much lower bandwidth and resource
usage, our method makes itself approachable for robots with limited
resources like UAVs and swarm robots. Our approach is able
a compress RGBD scenes of ~1MB in real-time to smaller plane clouds that typically require ~10KB. This makes these representations more suitable for large-scale real-time
mapping.

\begin{figure*}
\centering \subfigure[RGB image of a scene]{ %
\begin{minipage}[t]{0.3\textwidth}%
\centering %
\fbox{\includegraphics[width=1.9in,height=1.5in]{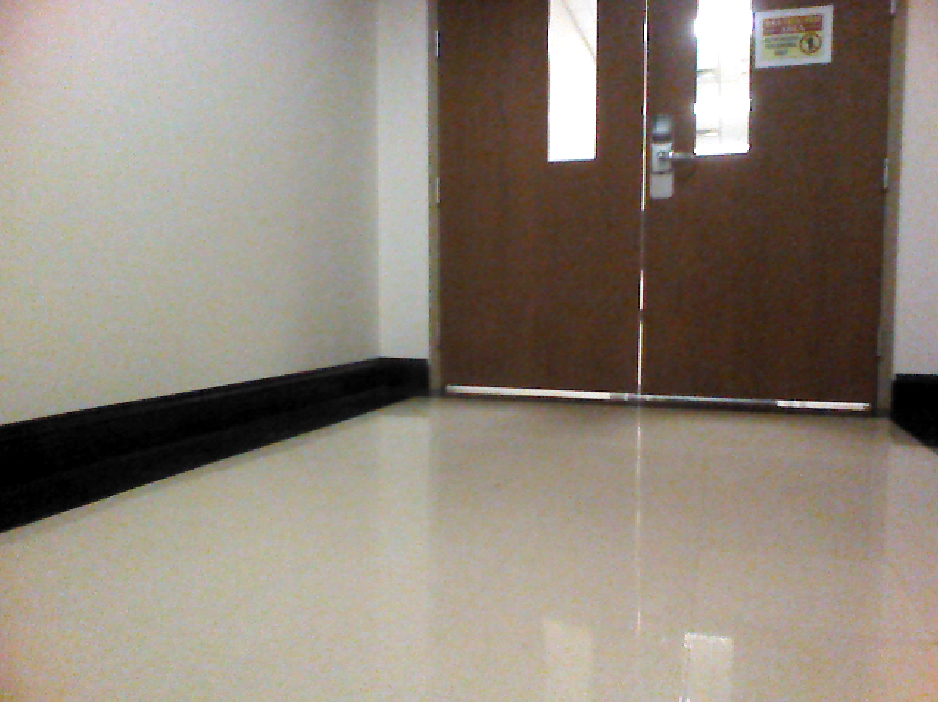}} %
\end{minipage}} \subfigure[Dense point cloud of the scene]{ %
\begin{minipage}[t]{0.3\textwidth}%
\centering %
\fbox{\includegraphics[width=1.9in,height=1.5in]{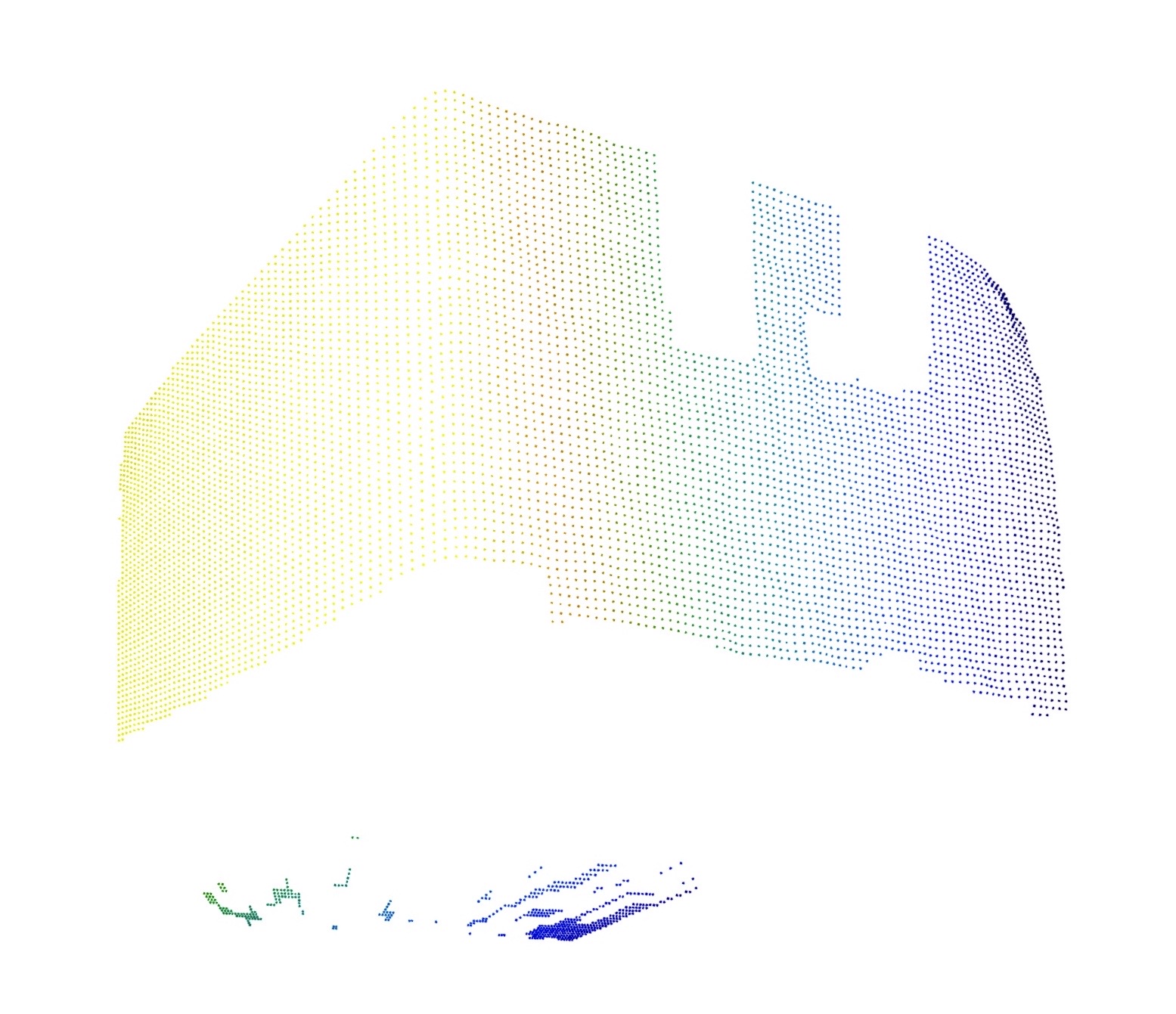}} %
\end{minipage}} \subfigure[Planar representation of the scene]{ 
\begin{minipage}[t]{0.3\textwidth}%
\centering %
\fbox{\includegraphics[width=1.9in,height=1.5in]{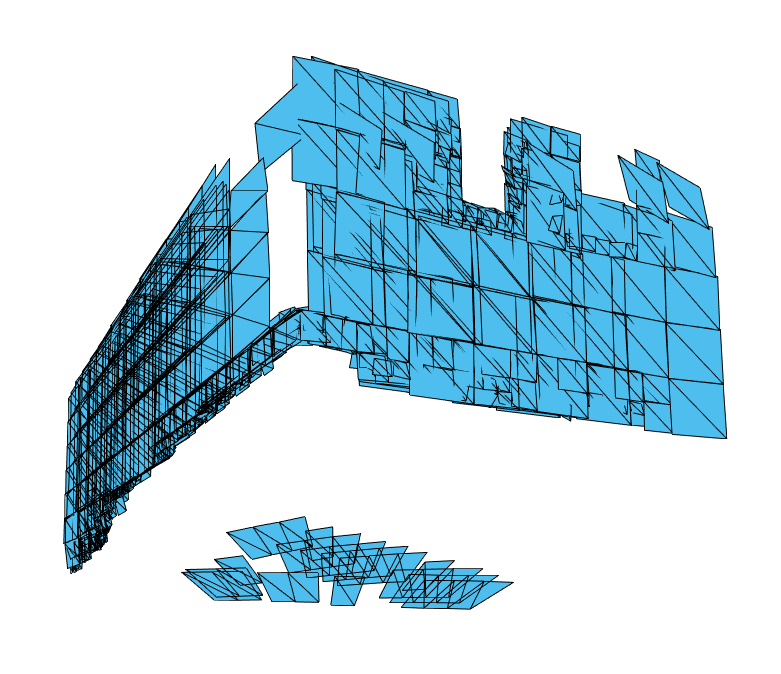}} %
\end{minipage}} \caption{\label{fig:building}The proposed SLAM approach compresses RGBD input
and converts the image data (left) and point cloud data (middle) to
a collection of planes (right) (textures omitted).  Advancements over state-of-the-art are provided via 3D point cloud compression.
 Adaptive 3D geometric compression of dense sensor data allows hosts heretofore
new capabilities to tailor the geometric detail needed for both local and
 over-network mapping applications.}
\end{figure*}

The key enabling component of our approach integration of a compressed map representation to the RGBD SLAM program. This provides a new representation for point cloud that can be quickly
calculated and represents the data to a similar degree of geometric
accuracy using far fewer parameters. This allows large collections of point measurements to be substituted with a small collection of local planar fits. Fig.\ref{fig:building} depicts the difference in the two representations. Fig.\ref{fig:building}(a) and (b) show a
close-up scene and the dense point cloud representation which is used
by many state-of-the-art SLAM work. Fig.\ref{fig:building}(c) shows
the sparse reconstruction outcome of our SLAM system in which the
limited size of planar patches are transformed from dense 3D surface
data.

\section{\label{sec:related}Related Work}

Our review of closely related research focuses on four areas: (1) SLAM solutions that address SLAM computational cost, (2) SLAM methods that incorporate plane models, (3) over-network SLAM solutions and (4) DVO-SLAM.

\subsection{Computational Cost}

One major challenge for current 3D SLAM approaches is to overcome the burden that SLAM and GNC using RGBD sensors places on the host's available computational resources. Centralized
approaches\cite{bailey2011decentralised,lazaro2013multi,dong2015distributed},
address computational cost by aggregating data from multiple robots at a central server having more computational resources where SLAM estimates can be calculated. Yet, such approaches are not viable for RGBD data since sharing this data requires prohibitively large network bandwidth. Further, this computational model does not scale for as the number of robots increase. A second approach solves the SLAM optimization problem via distributed computation approaches. In this context, robots utilize only local computation and communication to optimize the SLAM pose graph and estimate robot trajectories and environmental maps. Another distributed mapping algorithm\cite{choudhary2017distributed} optimizes the SLAM algorithm by sharing key informative features.  Recent research \cite{cieslewski2018data,wang2019active} have extended this implementation as a backend to larger SLAM solutions as a method to reduce the computational burden of solving  multiple robot trajectories in multi-robot systems. While these solutions can be data-efficient,  they are not applicable to RGBD 3D mapping scenarios. Further, these approaches lack representations required to approach 3D map building in bandwidth constrained contexts.

\subsection{Planar Shape Models for SLAM
}
There has also been significant interest in  probabilistic mathematical methods to
model trajectory and geometric shape to mitigate computational issues. Some work
\cite{gtsam,Cunningham13icra,5509154} succeeds in mitigating the propagation
of SLAM uncertainties and controlling the computational complexity
using probabilistic graphical models \cite{Kschischang:2006:FGS:2263225.2266465}. Compact shape models are also used for efficient surface data representation that can be associated and integrated with low computational complexity. These plane-involved SLAM solutions, however, either extends feature-based SLAM with the use of planar surfaces\cite{6225287}, or has an orthogonal assumption on the environment which is less applicable\cite{Nguyen06orthogonalslam:}. Dense planar SLAM\cite{6948422} uses a dense ICP method to estimate sensor poses which requires GPU for real-time computation. A quaternion-based minimal plane representation \cite{7139837} is utilized to update the planes during optimization without using GPU but the system does not perform in real-time. A real-time CPU only execution of dense planar SLAM algorithm\cite{hsiao2017keyframe} succeeds in exceeding current popular online 3D reconstruction methods in pose estimation while the computational cost can be further saved by aligning planes for loop closures instead of searching for 3D point pairs. Moreover, most of these planar SLAM systems are not distributed.

\subsection{Bandwidth}

Collaborative mapping in multi-agent environments require each agent
to share information about their past and current state. Due to the
bandwidth constraints and limited communication range, it is challenging
to share raw data or other large amounts of data among the agents
in a distributed SLAM system \cite{lajoie2019door}. A distributed communication network \cite{montijano2013distributed} is proposed where every robot only exchanges the local matches with its neighbors. The algorithm propagates local to global contexts to solve for a global correspondence. Some work \cite{nettleton2003decentralised,tardioli2015visual} manages to reduce bandwidth requirements by transmitting the subsets of the map information as a collection of sparse features. These sparse representations give rise to sparse map data that may contain large holes and ambiguous regions.

A proposed multi-agent system reduces the required communication bandwidth and the complexity by partitioning point clouds into parts and then compactly describing each part using discriminating features to efficiently
represent the environment\cite{dube2017online}. Some work\cite{cieslewski2017efficient,cieslewski2017efficient2,cieslewski2018data} minimizes
bandwidth usage by running place recognition on image frames and only sending the extracted place feature vectors to the robots. A resource-adaptive framework\cite{tian2018near,tian2019resource} is proposed for distributed loop closure detection by seeking to maximize task-oriented objectives subject to a budget constraint on total data transmission in front-ends.  While all of these SLAM systems work efficiently in reducing bandwidth, none of these SLAM solutions are designed for RGBD data nor do they provide dense 3D geometric surface maps as part of their SLAM solution.

\subsection{DVO SLAM}

Much recent work\cite{dtam,lsd,engel2016dso,svo} in real-time RGBD
sensor based visual SLAM involve direct methods where input data is
used directly with minimal processing and maps are dense. DVO SLAM,
originally proposed by Kerl et al.\cite{Kerl2013} in 2013, is still
in popular use today. It benefits from low drift odometry via minimization
of both geometric and photometric re-projection error between RGBD
image pairs. The odometry is made robust by the incorporation of motion
priors and the weighting of errors. As many 3D RGBD odometry algorithms
do, the map in this approach is modeled by a pose graph, where every
vertex is a camera pose with associated keyframe (intensity + depth),
and edges represent the relative Euclidean transformations between keyframe poses.
New keyframes are generated based on an entropy metric, and a similar
metric is used to identify loop closures. For DVO SLAM the map is
taken as the RGBD keyframe point clouds whose poses are determined by their current
\char`\"{}best estimate\char`\"{} in the optimized SLAM camera trajectory.  While this method is highly accurate there are several limitations including:
(1) the algorithm must be run on a single host, (2) other robots cannot benefit from estimated SLAM map data and (3) if the DVO map data were available, it would be prohibitive in size to share even over high-bandwidth network connections\cite{6631318,hsiao2017keyframe,7139837}. 

This work represents a significantly distinct point in the landscapae of SLAM algorithms. While the proposed approach is derived from DVO SLAM\cite{Kerl2013}, extensive modifications to the programmatic structure, visual odometry tracking and SLAM optimization set this work apart. In combination, these modifications create a high-performance graph-based distributed SLAM system that can dynamically adapt to the changing computational resources and bandwidth conditions available to the host computer. This is accomplished by separating the DVO SLAM algorithm into independent and asynchronous \textbf{\emph{frontend}} and \textbf{\emph{backend}} applications. These applications can run on the same or different CPUs or on completely distinct hosts connected over a low-bandwidth network. The \textbf{\emph{frontend}} is responsible for camera tracking over small changes in view and geographic regions. The \textbf{\emph{backend}} which receives compressed 3D geometric data from one or more frontends and integrates these data to construct maps. By operating directly upon compact representations of the RGBD data the computational and bandwidth resources consumed by processing and sharing map data is significantly mitigated.

\section{Methodology}

The goal of the proposed approach is to allow hosts to "fit" a version of DVO SLAM to their context. This enables robots with limited resources to contribute to and potentially benefit from 3D maps in a distributed 3D SLAM system. To achieve this goal, we introduce significant modifications to DVO SLAM\cite{Kerl2013} to produce a new SLAM algorithm that dynamically adapts to satisfy computational and bandwidth constraints set by the host computer. This goal is achieve by the following (4) modifications to the DVO-SLAM algorithm:
\begin{itemize}
\item Computational isolation of frontend tasks and backend tasks into separate execution units. 
\item Replacement of 3D point clouds with plane clouds in the graph-SLAM keyframe data.
\item A RGBD point cloud to plane cloud compression algorithm.
\item Optimization methods for plane clouds for delta pose and loop closure estimation. 
\end{itemize}

The cumulative effect of these modifications creates a completely new SLAM implementation for 3D map building from RGBD sensor data.

\subsection{Computational isolation of frontend tasks and backend tasks}

Our approach starts with the state-of-the-art implementation of a multi-threaded single-CPU
implementation of graph-SLAM as provided by DVO SLAM \cite{Kerl2013}.
As implied by its name, it uses Direct Visual Odometry to track camera pose changes from sensed RGBD image data by minimizing both intensity and depth reprojection error between RGBD image pairs, resulting in estimates of camera movement and associated uncertainty. DVO SLAM implements a keyframe approach to 3D RGBD mapping. Here, the 3D map is taken as a collection of sensed 3D keyframes placed at the estimated pose of the RGBD camera when the keyframe was sensed. DVO SLAM solves the mapping problem using two interconnected mapping components: (1) a local map consisting of a single keyframe that is updated by live RGBD sensor data and (2) a global map which stores all other keyframes historically recorded to represent the history of the RGBD camera trajectory.

Our definition of the new SLAM system frontend consists of the DVO camera tracking algorithm and local map component of the SLAM mapping which only keeps the most recently measure keyframe. This allows the frontend to construct maps of small geographic regions without a need to communicate this information to the backend. Once the local map is "completed" due to a significant (tunable) change in viewpoint, e.g., rotation, or a change in position, i.e., translation, the local map is serialized and communicated across the network (or host) to the backend application and a new local map is created. 

Our definition of the new SLAM system backend holds the global map data from DVO SLAM. The backend is responsible for potentially large scale map integration. Map integration is accomplished by embedding the graph-SLAM problem into the generic g\textsuperscript{2}o graph optimization library \cite{5979949}. G\textsuperscript{2}o graph construction crates graph vertices from estimated camera poses and their uncertainty and attributes each vertex with keyframe (plane cloud) data. Edges connect vertices and imply that the sensor data from the connected keyframes are constrained, i.e., the two keyframe include overlapping views/measurement of the same underlying geographic map regions. The backend can attempt to create new edges, i.e., loop closures, by attempting to detect instances of this circumstance. In DVO SLAM loop closures are found by speculative matching using spatial locality  as a guide for which keyframe pairs are candidate matches.

The resulting isolated frontend and backend applications allow completely new modular SLAM deployments. For example, robots having RGBD sensors and low computational resources might run only a frontend or perhaps only a backend. Further separate processors on a host might independently be tasked with these tasks allowing new energy saving strategies. For our application, we wish to perform centralized real-time mapping on a desktop host in near real-time as one or more remote vehicles contribute map data. Note this modularity also allows for decentralized mapping by endowing each robot in a multi-robot network with a backend.

\subsection{\label{subsec:Graph-SLAM}Graph SLAM}

The graph SLAM approach to solving the SLAM problem uses a graph model to capture state evolution over time. It is typically more efficient in space and time than other approaches because robot states are often described by a sparse collection of locally correlated states (poses). As mentioned previously, graph nodes are robot poses and edges are relative Euclidean transformations that connect poses.
Optimization of the graph estimates the \textit{full} trajectory of the agent, i.e.
the sequence of poses $\mathbf{x}_{1:T},$ that minimizes errors in the observed constraints. Constraints come in the form of pairwise correspondences, i.e., matches between observed map keyframe data, $\mathbf{m}$, and realize as edges in the graph-SLAM model. This is also known as the \textit{full} SLAM problem, which estimates all previous states of the agent trajectory. This
is in contrast to filtering based approaches, which seek to solve
the incremental or on-line SLAM problem, whereby previous state estimates
are marginalized out.

As the agent moves through the unknown scene it generates vertices as keyframes and sequential keyframes are always connected by odometry edges, $\mathbf{u}_{1:T}$, that estimate of the relative pose from one keyframe to the next. Non sequential keyframes may also be connected by these edges when the camera rotates in circles or loiters in an area. 

Each inclusion of a new edge changes the globally optimal solution which typically effects poses close to the new observation more than geographically distant locations. Loop closures between distant nodes (via edge traversal) are particularly valuable for controlling the error and uncertainty in map estimation and new methods for remote loop closure is a topic of great interest for accurate SLAM map building. 

Finally, the graph-SLAM problem also permits the inclusion of features or landmarks in the environment as constraints. Here salient aspects of observed data denoted,  $\mathbf{z}_{1:T}$, are used to define new nodes with their own metrics for value and uncertainty. 

The full SLAM problem is then formulated as the estimation of the
posterior probability of the trajectory, $\mathbf{x}_{1:T},$ and
the environmental map, $\mathbf{m}$, given a sequence of motion estimates,
$\mathbf{u}_{1:T}$, landmark measurements, $\mathbf{z}_{1:T}$, and
the initial pose $x_{o}$ as in Equation \eqref{eq:Slam-Posterior}.

\begin{equation}
p(\mathbf{x}_{1:T},\mathbf{m}|\mathbf{u}_{1:T},\mathbf{z}_{1:T},x_{o})\label{eq:Slam-Posterior}
\end{equation}

Modeling the full SLAM problem in a graphical way leads to an optimization
problem over a sum of nonlinear quadratic constraints in the graph.
These constraints are built up over time as the agent explores the
environment in the form of both motion and measurement constraints.
In this formulation, agent and landmark poses correspond to graph
vertices, while the edges represent spatial constraints (transformations)
between these vertices as a Gaussian distribution.

Motion estimates form a constraint directly between agent poses, while
landmark measurements form constraints between agent poses through
the mutual observation of a landmark. Individual measurements therefore
link poses to the landmarks in the graph, and during the optimization
these measurements are then mapped to constraints between poses that
observed the same landmark at different points in time.

Once the graph is constructed, we seek the configuration of the graph
vertices that best satisfy the set of constraints. That is, we seek
a Gaussian approximation of the posterior distribution of Equation
\eqref{eq:Slam-Posterior}. The optimal trajectory, $\mathbf{x}^{\star}$,
is estimated by minimizing over the joint log-likelihood of all constraints
as described by Equation \eqref{eq:Slam-Minimization}. \cite{Thrun:2005:PR:1121596}

\begin{equation}
\mathbf{x}^{\star}=\min_{\mathbf{x}}\left(x_{o}^{T}\Omega_{o}x_{o}+\sum_{t}[x_{t}-g(u_{t},x_{t-1})]^{T}R_{t}^{-1}[x_{t}-g(u_{t},x_{t-1})]+\sum_{t}[z_{t}-h(x_{t},m_{i})]^{T}Q_{t}^{-1}[z_{t}-h(x_{t},m_{i})]\right)\label{eq:Slam-Minimization}
\end{equation}

The leftmost term, $x_{o}^{T}\Omega_{o}x_{o}$, represents our prior
knowledge of the initial pose. Often, $x_{o}$, is set to zero, anchoring
the beginning of the trajectory to the global coordinate system origin.

The middle term describes a sum over the motion constraints, where
$g(u_{t},x_{t-1})$ is a function that computes the pose resulting
from applying the odometry estimate obtained at time $t$, $u_{t}$,
to the pose $x_{t-1}$. The residual vector, $x_{t}-g(u_{t},x_{t-1})$,
is then the difference between the expected pose, $x_{t}$, and the
pose observed by applying the motion estimate. The matrix $R_{t}^{-1}$
is the information matrix or inverse covariance matrix associated
with the motion constraint at time $t$. This matrix encodes the precision
or certainty of the measurement, and serves to weight the constraint
appropriately in the optimization.

The rightmost term describes a sum over landmark constraints. The
function $h(x_{t},m_{i})$ computes a global pose estimate of a given
landmark, $m_{i}$, using a local measurement of the landmark and
the pose of the agent, $x_{t}$, when the landmark was observed. The
residual $z_{t}-h(x_{t},m_{i})$, is then the difference between the
expected landmark pose in the global coordinate system, and the estimated
global landmark pose resulting from the local landmark measurement.
As before, the matrix $Q_{t}^{-1}$ is the information matrix or inverse
covariance matrix associated with the landmark constraint at time
$t$.

Minimization of this error function is a nonlinear least squares problem,
and can certainly be solved using nonlinear optimization approaches,
e.g., Levenberg-Marquardt, Gauss Newton, etc. It can also be solved
efficiently by exploiting the sparse structure of the graph SLAM formulation
\cite{5681215}, thereby allowing the use of sparse methods such as
sparse Cholesky decomposition or the method of conjugate gradient.
Many graph optimization libraries, such as the g\textsuperscript{2}o
library \cite{5979949}, leverage these sparse methods in order to
quickly optimize large graphs.

DVO SLAM's formulation of graph-SLAM incorporates only motion constraints into the optimization (the third term in equation (\ref{eq:Slam-Minimization}) is 0). Motion constraints originate from two sources: (1) appending/integrating new local map keyframes into the global map from the frontend and (2) detecting loop closures by speculatively evaluating correspondence hypotheses between unconnected keyframe pairs. In both cases, the function $g(u_{t},x_{t-1})$ from equation (\ref{eq:Slam-Minimization}) produces a pairwise motion estimate for the edge-connected keyframe pair by computing the relative motion that minimizes the corresponding keyframe data error. The function $g(u_{t},x_{t-1})$ in this case is the DVO odometry algorithm which, when invoked on the on the RGBD data stored in the keyframe pair, provides motion optimization error estimates.

Similar to DVO SLAM, our extension performs graph-SLAM incorporating only motion constraints into the optimization. Motion constraints also originate from the same two sources: (1) appending/integrating new local map keyframes into the global map from the frontend and (2) detecting loop closures by speculatively evaluating correspondence hypotheses between unconnected keyframe pairs. However, in contrast to DVO SLAM, our backend keyframes store compressed plane clouds in lieu of RGBD data. This fundamentally changes the function $g(u_{t},x_{t-1})$ in the second term of the graph-SLAM optimization problem in equation (\ref{eq:Slam-Minimization}) from an RGBD point cloud alignment problem to that of aligning compressed plane clouds. The new optimal pairwise motion estimate for a keyframe pair replaced by a new odometry algorithm $g(u_{t},x_{t-1})$ that estimates the relative motion of a pair of point clouds by  minimizes the corresponding pairwise plane cloud data error. The plane cloud odometry algorithm applied is described in \S \ref{Compressed Plane Cloud Odometry}.

\subsection{\label{subsubsec: planar representation}A RGBD point cloud to plane cloud compression algorithm}

We seek to compress the sensed RGBD imaged substituting 3D planar regions for collections of locally flat depth measurements. This approach for 3D scene compression promises to significantly reduce the size of the RGBD image data which serves to reduce both memory usage and computational costs.

RGBD point clouds images are converted to plane clouds using a recursive plane fitting algorithm. The algorithm consists of three conceptual steps:
\begin{enumerate}
\item Tile the image using rectangles from a quad-tree decomposition and add them to the plane fitting queue.
\item Remove the topmost tile data in the plane fitting queue and fit a 3D plane to the data. 
\begin{enumerate}
\item If the fit is inaccurate, subdivide the tile and add the 4 new sub-tiles to the plane fitting queue.
\item If the fit is accurate, add the computed plane coefficients to the plane cloud.
\end{enumerate}
\item Continue iterating until either the bandwidth or computational time constraints are exceeded or all tiles have been processed. 
\end{enumerate}

A more detailed version of this algorithm is detailed in Algorithm \ref{alg:fitting}. 

Each plane fit is accomplished very quickly using an ultra fast 
least squares plane fitting formulation\cite{7925286} for RGBD sensors that uses the RGBD camera calibration parameters to fit planes directly to the measured depth image without reconstruction of 3D $(x,y,z)$ point cloud coordinates.

This accelerated fitting of planes is made possible by a rearrangement of the
standard plane fitting error functions for RGBD data. While standard
planar representations adopt a form of equation $aX+bY+cZ+d=0$, we
substitute the RGBD 3D reconstruction equations described in the
previous work \cite{7925286} for the variables $X$ and $Y$ then
simplify the resulting equation to fit directly to only the measured RGBD depths and, in doing so, save significant computational cost.

The re-arrangement of the terms gives $aZ\tan\theta_{x}+bZ\tan\theta_{y}+cZ+d=0$ where $\tan\theta_{x}=(x+\delta_{x}-c_{x})/f_{x}$ and $\tan\theta_{y}=(y+\delta_{y}-c_{y})/f_{y}$  are values that can be pre-computed from the known integer $(x,y)$ pixel coordinates and the calibration parameters. Multiplying this equation by $1/Z$ gives us the equation \eqref{eq:RGBD_space_plane_eq}.

\begin{equation}
a\tan\theta_{x}+b\tan\theta_{y}+c+\frac{d}{Z}=0\label{eq:RGBD_space_plane_eq}
\end{equation}

In \eqref{eq:RGBD_space_plane_eq} we notice the coefficients $a,b,c$ are functions of \emph{only the  camera calibration parameters} which determine, for each $(x,y)$ pixel location, the
values of $\tan\theta_{x}$ and $\tan\theta_{y}$. As a result, the plane coefficients and specifically, \emph{only} plane coefficient $d$ is a linear function of the measured depth.

For implicit plane fitting using the re-organized equation \eqref{eq:RGBD_space_plane_eq},
the new monomial vector becomes $\mathbf{{M}}_{i}=[\begin{array}{cccc}
\tan\theta_{x_{i}} & \tan\theta_{y_{i}} & 1 & \frac{1}{Z_{i}}\end{array}]$ and the scatter matrix has elements:

\begin{equation}
\mathbf{M}^{t}\mathbf{M}=\sum_{i=1}^{N}\left[\begin{array}{cccc}
\tan\theta_{x_{i}}{}^{2}\\
\tan\theta_{x_{i}}\tan\theta_{y_{i}} & \tan\theta_{y_{i}}^{2}\\
\tan\theta_{x_{i}} & \tan\theta_{y_{i}} & 1\\
\frac{\tan\theta_{x_{i}}}{Z_{i}} & \frac{\tan\theta_{y_{i}}}{Z_{i}} & \frac{1}{Z_{i}} & \frac{1}{Z_{i}^{2}}
\end{array}\right]\label{eq:tan_theta_scattermatrix}
\end{equation}

where the symmetric elements of the upper-triangular matrix have been
omitted to preserve space. It is important to note that only 4 elements
of this matrix, $[\begin{array}{cccc}
\frac{\tan\theta_{x_{i}}}{Z_{i}} & \frac{\tan\theta_{y_{i}}}{Z_{i}} & \frac{1}{Z_{i}} & \frac{1}{Z_{i}^{2}}\end{array}]$, depend on the measured depth data and, as such, this matrix requires
less (\textasciitilde{}50\% less) operations to compute. The upper
left 3x3 matrix of $\mathbf{M}^{t}\mathbf{M}$ \emph{does not depend
upon the measured sensor data}. As such, once the calibration parameters
of the camera are known, i.e., $(f_{x},f_{y})$, $(c_{x},c_{x})$,
$(\delta_{x},\delta_{y})$, many of the elements of $\mathbf{M}^{t}\mathbf{M}$
are determined and \emph{may be pre-computed before any data is measured}
with the exception of the 4 elements in the bottom row. The plane coefficients of a fit can be found as the eigenvector, $\mathbf{v}_i=[a,b,c,d]^t$, corresponding to the smallest eigenvalue, $\lambda_i$, of the matrix  $\mathbf{M}^{t}\mathbf{M}$.

\begin{algorithm}[ht]
\SetAlgoLined \textbf{Input:} RGBD image $I$, initial tile size
$B^{2}$ (unit: pixels), minimal tile size $m^{2}$, time budget $T$
(unit: s)\\
 \textbf{Output:} planar surface fits $P$\\
 \textbf{Parameters:} $\epsilon$ - error threshold, current fitting
size $\beta^{2}$, tile set $S$ \\
 $\beta=B$ \;

$S.initialize(I)$\; // initialize the input image as fitting candidate
\\
 \While{$timeConsumption<T$ $\&\&$ $!S.empty()$}{ \If{$\beta$
exists}{ \ForAll{$quad_{j}\in S$.decomposite($\beta$)} { //
decomposite subdivided set into $\beta\times\beta$ blocks\\
 \eIf{checkTileDepth($quad_{j}$)} { // check existence of depth
values \\
 $\pi_{j}=quad_{j}.planefitting$\; // one planar fit $\pi_{i}$
\\
 \eIf{$\pi_{j}.err<\epsilon*avgDepth(quad_{j}.corners)$} { \eIf{(planarityCheck)}
{ // if the plane fits a real surface \\
 $P.add(\pi_{j})$\; \hspace{0.1cm} // one fit is accepted }{
\If{$\pi_{j}.area()>m^{2}<<2$}{ $S.add(\pi_{j}.subdivide())$\;
// subdivide planar fits into 4 equally smaller elements } }}{
\If{$\pi_{j}.area()>m^{2}<<2$}{ $S.add(\pi_{j}.subdivide())$\;
} } }{ \If{$\pi_{j}.area()>m^{2}<<2$}{ $S.add(\pi_{j}.subdivide())$\;
} } }} $\beta=\beta>>1$ \hspace{0.1cm} // shrink the fitting
block size } return $P$\; \caption{\label{alg:fitting}Fit planes to detected surface}
\end{algorithm}

The implementation of our planar regions extraction is shown in Algorithm
\ref{alg:fitting}. Some considerations are: 
\begin{itemize}
\item The initial RGBD image is covered by an $B\times B$ array of tiles (tile size of 80x60 pixels). This size is important and implicitly defined the size of the largest planar surfaces we  expect to observe in RGBD image data. Incorrect initial sizes waste computation by either requiring subdivision (too big) or needlessly splitting planes (too small). 
\item Before fitting, a quick check on the tile depth is applied to ensure
the existence of the depth values at four corners of the tile. Avoid fitting plane models to tiles which may have significant portions of missing depth data.
\item Adaptive subdivision using the Quadtree\cite{Quadtree} model seeks to cover all the depth data using planar surfaces. Thresholds use the averge fitting error for the tile to determine which planes are accepted. Thresholds change based on the depth to accommodate for the depth noise model of RGBD camera. This model indicates that error at maximum range ($~$7m) can be nearly an order of magnitude larger than those at short range ($~$0.5m).

\item Before fitting, a planarity check to establish the likelihood that a plane will fit the data. The planarity check uses depth at three of the four corners to define a plane and then evaluates the Euclidean distance of the fourth corner to this plane. A threshold on the error determines if the tile data is subdivided (large error) or if a plane fit is attempted (small error).
\end{itemize}
It is also noted that our execution of Algorithm \ref{alg:fitting}
is breadth-first. It first detects and fits all of the valid surface
area with planar tiles at current subdivision level then if possible,
applies detection and fitting to all of the next subdivided elements.
Planar map, as a compact representation, provides significant resource
savings while a point-based representation ignores the scene geometry
hence suffers from considerable redundancy. With all planes being
approximated, it is essential to have an accurate and a concise representation
of the scene, that is, a rate-distortion optimal 3D scene structure,
which is estimated by a joint optimization of the number of bits/points
used to encode the extracted dense depth field, and the quality/accuracy
of representation rendered from the 3D structure.\cite{rate-distortion}

\subsection{\label{Compressed Plane Cloud Odometry}Compressed Plane Cloud Odometry Estimation and Loop Closure}

Each pairwise plane cloud hypothesis is evaluated using the Iterative Closest Algebraic Plane (ICaP) algorithm (Algorithm \ref{alg:iterative-closest-algebraic-plane}). This algorithm uses the compact representation of the world in terms of planar algebraic surfaces, i.e., surfaces having equation $ax+by+cz+d = 0$, to establish the likelihood that a given hypothesized plane cloud pair can be aligned with the intent to align the overlapping observed geometric map regions in a manner similar to puzzle-solving. This is accomplished by minimizing an error functional that solves for both the correspondence of planar surfaces between the maplets and the Euclidean transform that aligns these algebraic surfaces. The magnitude of the algebraic alignment error then serves as a goodness-of-fit metric to validate or refute plane pair hypotheses and as a covariance metric for the g\textsuperscript{2}o optimization library. 

\begin{algorithm}[h]
\SetAlgoLined
\hspace{0.1cm}\textbf{Input:} Two sets of plane coefficients $\pi_j,\pi_l$ and initial transform $\mathbf{T}_0$\\
\textbf{Assume:} size($\pi_j$) $>$ 3, size($\pi_l$) $>$ 3, size($\pi_j$) $\geq$ size($\pi_l$)\\
\hspace{0.1cm}\textbf{Output:} $\widehat{\mathbf{T}}$ the transform minimizing Equation (\ref{eq:lsq_plane_alignment})\\
\hspace{0.1cm}\textbf{Parameters:} $\tau_T$ - threshold for convergence\\
$\mathbf{T}_i = \mathbf{T}_0$\;
$\Delta \mathbf{T} = \infty$\;
N = size($\pi_l$);\hspace{1cm} // number of matches to compute\\
\While{$\Delta \mathbf{T} \geq \tau_T$}{
 $matchIndexPairs = \{\emptyset\}$ \hspace{0.1cm}// closest plane index-pairs\\
 \ForAll{${l_0} \in 1,2,...,N$} {
 $j_0 = \min_{j}\|\pi_{j}-(\mathbf{T}_{i}^{-1})^t\pi_{l_0}\|^2$; \\
 matchIndexPairs.add($\{l_{0},j_{0}\}$)\;
 } 
 // solve Equation (\ref{eq:tan_theta_scattermatrix}) as described in article \S \ref{subsubsec: planar representation}\\
 $\mathbf{T}_{i+1} = $alignPlanePairs(matchIndexPairs,$\pi_j,\pi_l$);\\
 $\Delta \mathbf{T} = \|\mathbf{T}_{i+1}-\mathbf{T}_{i}\|^2$;\\
 $\mathbf{T}_{i} = \mathbf{T}_{i+1}$\;
 }
return  $\mathbf{T}_{i}$\;
 \caption{\label{alg:iterative-closest-algebraic-plane}Iterative Closest Algebraic Plane (ICaP)}
\end{algorithm}

While the description provided in Algorithm \ref{alg:iterative-closest-algebraic-plane} describes the program flow, our mathematical solution for alignment of corresponding algebraic plane equations is described in the following paragraphs. These steps occur inside the \emph{alignPlanePairs()} function from Algorithm \ref{alg:iterative-closest-algebraic-plane} which includes the following statement:
\begin{equation}
\mathbf{T}_{i+1} = alignPlanePairs(matchIndexPairs,\pi_j,\pi_l);
\end{equation}

This function computes the optimal alignment (in the least squares sense) between planes that are hypothesized to have the same equation up to an unknown Euclidean transformation. For our derivation we denote $\pi_j$ and $\pi_l$ as two collections of $N$ plane equations. Equation (\ref{eq:lsq_plane_alignment}) expresses the optimization problem at hand. Here we seek to estimate the transformation $\widehat{\mathbf{T}}_{i \rightarrow j}$ that takes the planes of $\pi_l$ into the coordinate system of planes $\pi_j$. Note that Euclidean transformations, when applied to planes, follow the transformation rule $\pi^{'} = (\mathbf{T}^{-1})^t\pi$ for $\pi^t=\begin{bmatrix}a & b & c & d\end{bmatrix}$.
\begin{equation}
\label{eq:lsq_plane_alignment}
    \widehat{\mathbf{T}}_{i \rightarrow j} = \min_{\mathbf{T}_{i \rightarrow j}}\sum_{\{i,j\} pairs}	\|\pi_j-(\mathbf{T}_{i,j}^{-1})^t\pi_i\|^2 
\end{equation}

As in least squares point alignment \cite{Umeyama:1991:LET:105514.105525}, the solution is obtained by decomposing the problem into two steps: solve for the rotation,  $\widehat{\mathbf{R}}$, and solve for the translation, $\widehat{\mathbf{p}}$. After solving for these variables separately, the optimal transformation can be constructed as shown in Equation (\ref{eq:T_internals}).
\begin{equation}
\label{eq:T_internals}
\widehat{\mathbf{T}} = \begin{bmatrix}
            \widehat{\mathbf{R}} & \widehat{\mathbf{p}}\\ 
            \mathbf{0} & 1\\
        \end{bmatrix}
\end{equation}

We solve for the rotation, $\widehat{\mathbf{R}}$, that aligns the orientations of corresponding planes by finding the rotation that maximally aligns their normals, i.e., the vectors formed by the first three coefficients of each matching plane pair. To do so, we form a covariance matrix of the matching plane normal vectors as shown in Equation (\ref{eq:normal-covariance}).
\begin{equation}
\label{eq:normal-covariance}
C_{\mathbf{n}} = \sum_{\{i,j\} pairs} \mathbf{n}_i\mathbf{n}_j^t 
\end{equation}

We then decompose the covariance matrix using singular value decomposition, $svd()$, to compute the orthogonal transformation that minimizes the squared error, aligning corresponding plane normals as shown in Equation (\ref{eq:svd-decompostion}).
\begin{equation}
\label{eq:svd-decompostion}
\mathbf{U}\Lambda\mathbf{V}^t = svd(C_{\mathbf{n}})
\end{equation}

As pointed out in related literature on point alignment \cite{Umeyama:1991:LET:105514.105525}, the optimal rotation from Equation (\ref{eq:svd-decompostion}) could require a reflection. To find the optimal right-handed rotation, we must replace the smallest eigenvalue with $det(\mathbf{U}\mathbf{V}^t)$ to obtain the right handed rotation incurring smallest additional error beyond our original solution as shown in Equation (\ref{eq:rotation-reflection}).
\begin{equation}
\label{eq:rotation-reflection}
\widehat{\mathbf{R}} = \mathbf{U}\begin{bmatrix}
            1 & 0 & 0\\ 
            0 & 1 & 0\\ 
            0 & 0 & det(\mathbf{U}\mathbf{V}^t)\\
        \end{bmatrix}\mathbf{V}^t 
\end{equation}

\begin{figure*}
\centering \subfigure[Two plane clouds before alignment]{ %
\begin{minipage}[t]{0.45\textwidth}%
\centering %
\fbox{\includegraphics[width=2.2in,height=1.5in]{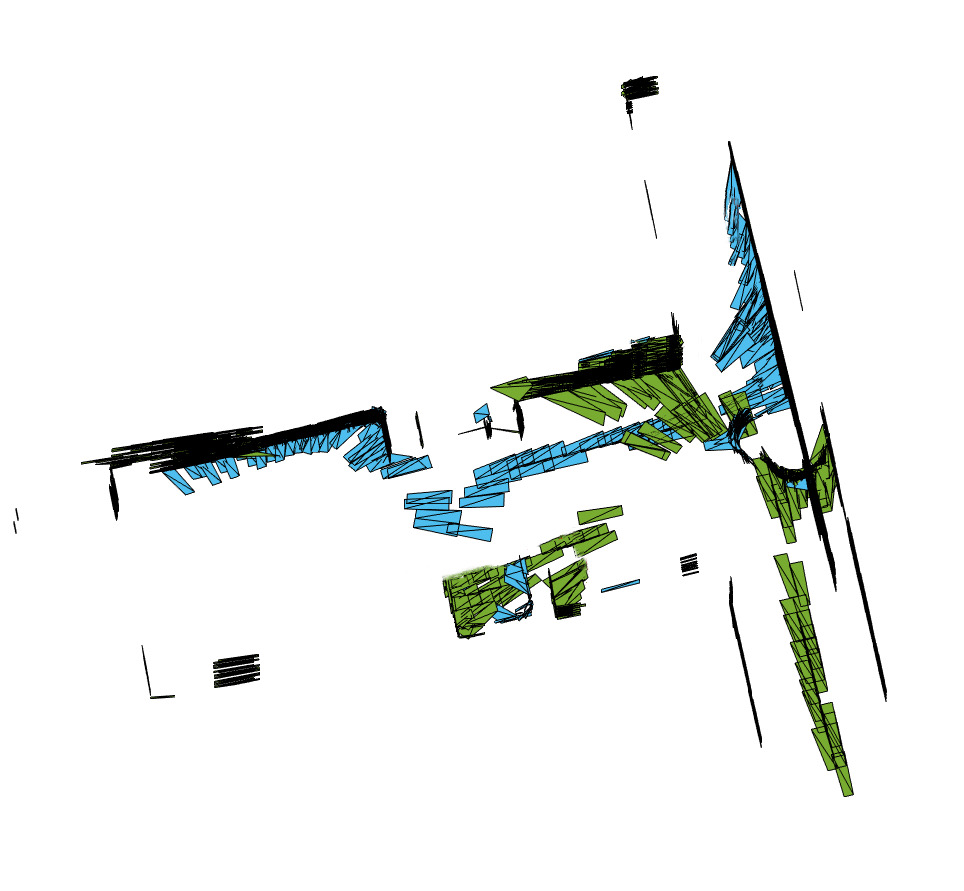}} %
\end{minipage}} \subfigure[Aligned plane clouds]{ 
\begin{minipage}[t]{0.45\textwidth}%
\centering %
\fbox{\includegraphics[width=2.2in,height=1.5in]{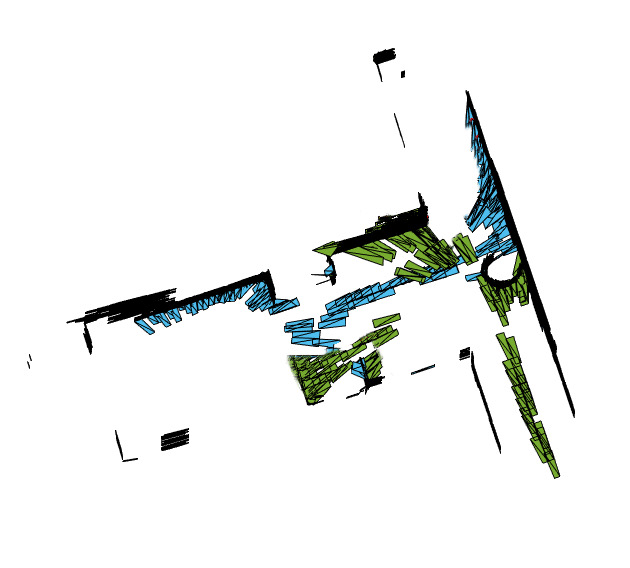}} %
\end{minipage}} \caption{Pairwise matches of plane clouds are matches quickly using an ICaP odometry/alignment algorithm. \label{fig:alignment}}
\end{figure*}

Using our solution for the optimal rotation, we then solve for the translation component of the transform that brings the two plane sets into final alignment. To do so, we form a matrix of the aligned normal vectors of the set $\pi_j$, $\widehat{\mathbf{N}}_j$. We also form vectors of the fourth coefficients of the corresponding planes to be aligned; $\mathbf{d_l}$ and $\mathbf{d_j}$. The mathematical representation for these variables are shown in the Equations (\ref{eg:vectors-for-translation-solution}).
\begin{equation}
\begin{matrix}
\label{eg:vectors-for-translation-solution}
\widehat{\mathbf{N}}_i^t = \widehat{\mathbf{R}}\begin{bmatrix}
            \mathbf{n}_{i_0} & \mathbf{n}_{i_1} & ... & \mathbf{n}_{i_N}
        \end{bmatrix}\\
        \mathbf{d_j}^t = 
        \begin{bmatrix}d_{j_0} & d_{j_1} & ... & d_{j_N} \end{bmatrix}\\
        \mathbf{d_i}^t = 
        \begin{bmatrix}d_{i_0} & d_{i_1} & ... & d_{i_N} \end{bmatrix}\\
        \end{matrix}
\end{equation}

Using these matrices and vectors we can write the explicit least squares solution for the translation component, $\widehat{\mathbf{p}}$, as shown in Equation (\ref{eq:solution-for-translation}).
\begin{equation}
\label{eq:solution-for-translation}
\widehat{\mathbf{p}} = ({\widehat{\mathbf{N}}_i^t}\widehat{\mathbf{N}}_i)^{-1}{\widehat{\mathbf{N}}_i}^t(\mathbf{d_i} - \mathbf{d_j})
\end{equation}

\section{Results}

Our experiments designate a path for TurtleBots to follow on the 2nd
floor of the EPIC building where the ECE Department and our laboratory
is located within the UNC Charlotte campus. The TurtleBots are equipped
with an Orbbec Astra pro RGBD camera (depth image: $640\times480$
@30fps) and our SLAM system is running on a Intel Core 2.3 GHz i5-6200U
laptop without GPU. We run our TurtleBots in five different hallways with the
length of at least 12 meters and generate a total of 1008 keyframes.
To be concise in this section, we randomly sample 100 keyframes and
focus on them. We average the results from these images for analysis
in this section.

Our first analysis examines the memory or, equivalently, bandwidth resources needed for the 
system to build a ''well-described'' map and how much data is required for this compressed representation. A well-described map should contain enough geometric
information (e.g. point cloud, planes and etc.) to represent a scene.
Ultimately, we seek to balance between the computational cost and
the bandwidth needed.

\subsection{Bandwidth}

Intuitively, one advantage of planes over point cloud when explaining
a scene is the reduction on the size of map data transmitted among
agents. Table \ref{tab:message size} shows the bandwidth required
for the data streams of planar and point cloud implementations. In
our approach, planes are in the Hessian normal form and thus it takes
as few as 8 float variables to describe a finite plane while some
redundancy reduction may be found. Our experiments show that the naive
implementations would require approximately 0.044MB per sensed image,
or 1.2MB of data for the dense point cloud, and while some saving
could be found, it is clear the planar representation is significantly
more compact. Note that RGB data is excluded here because we only
consider geometric information. The result shows using planar map
representation requires about 27 times less data to be transmitted
than the raw point cloud representation otherwise would. When communications
are required highly frequently, the ability to share more data per
unit time becomes increasingly appealing.

\begin{table}[h]
\noindent \centering{}%
\begin{tabular}{lcccc}
\hline 
 & mean  & min  & max  & std dev\tabularnewline
\hline 
\hline 
depth images (MB/image)  & 1.20  & 1.20  & 1.20  & 0.00\tabularnewline
\hline 
Planar maps (MB/image)  & 0.044  & 0.036  & 0.050  & 0.004 \tabularnewline
\hline 
\end{tabular}\vspace{1.5mm}
 \caption{Bandwidth of local map using planar representations shows significant
reduction compared to point cloud.}
\label{tab:message size} 
\end{table}

\subsection{Trade-off and Rate-Distortion}

\begin{figure*}[h]
\centering \subfigure[Trade-off between scene area fitted and bandwidth]{
\begin{minipage}[t]{0.45\textwidth}%
\centering \includegraphics[width=3in,height=2.5in]{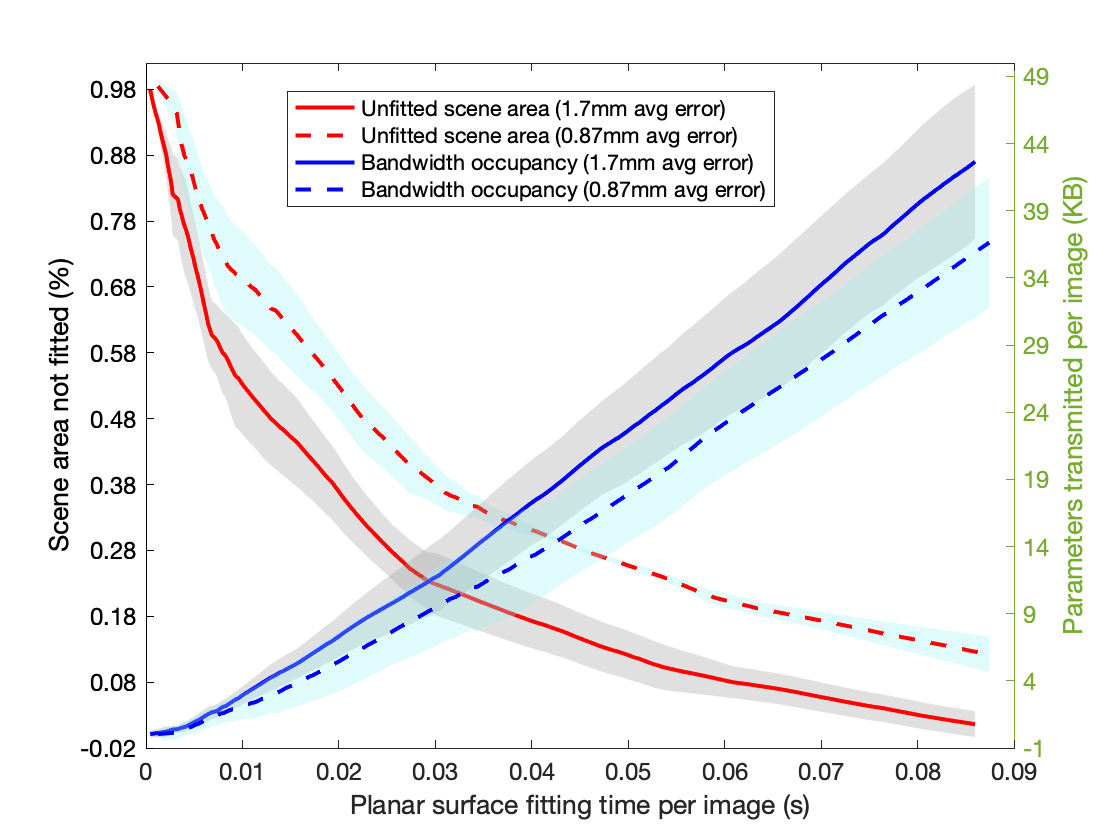}
\label{subfig:tradeoff1} %
\end{minipage}} \subfigure[Plane fits extracted from per image]{ 
\begin{minipage}[t]{0.45\textwidth}%
\centering \includegraphics[width=3in,height=2.5in]{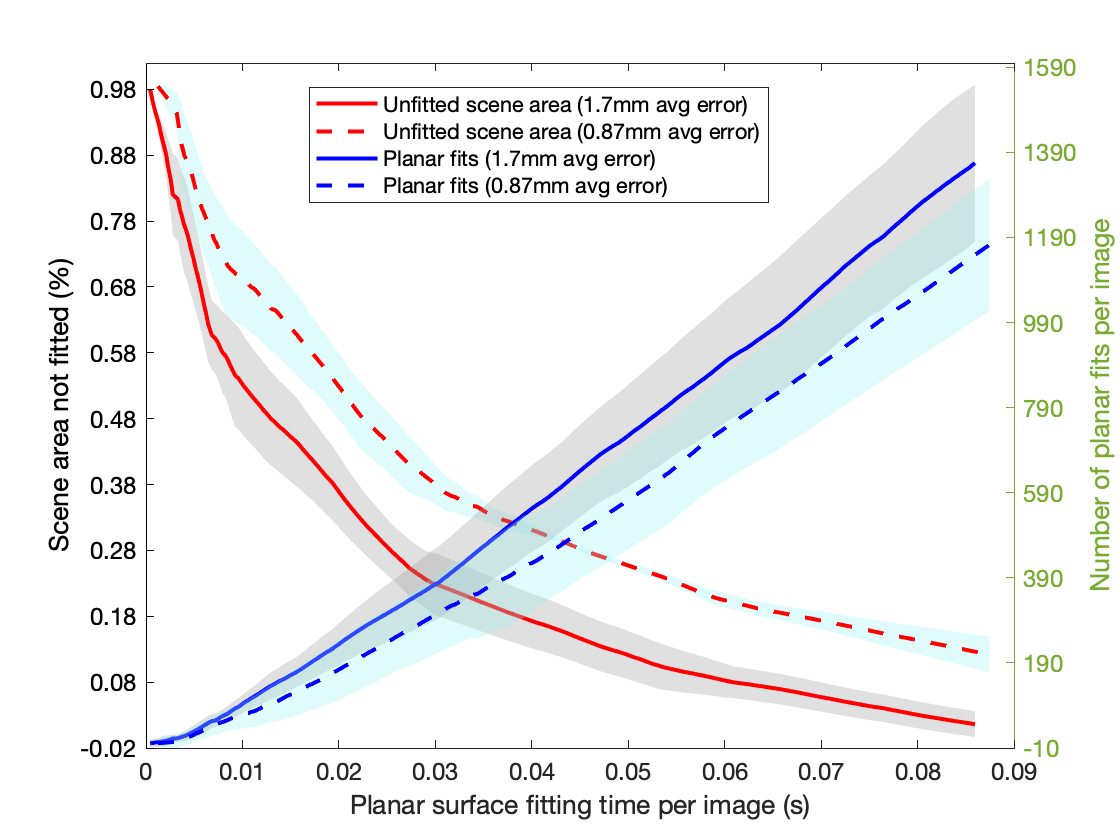}
\label{subfig:tradeoff2} %
\end{minipage}} \caption{Adaptive geometric sensitivities of our graph-SLAM system.}
\end{figure*}

In our experiments, the computation cost is measured in terms of plane
detection and fitting time. We measure the ability of the system to
explain a scene by the percentage of surface area (pixels) in
the scene that is fitted by planar blocks. In the mean time, we compare
the bandwidth required for sharing the planar map data. We set the
time budget 80$ms$, initial tile size $24\times24$ pixels and minimal
tile size $8\times8$ pixels. Fig.\ref{subfig:tradeoff1} shows the
trade-off relationship given a fitting error tolerance. The tolerance
is the maximal allowable fitting error. We calculate the average point-wise
distance to the planar fit. A fit is accepted when its error is below
the threshold. As we can see, for the solid curve (error tolerance
13.1$mm$), the surface area in the scene can be nearly 100\% fitted
within given time budget, while for the dash curve (error tolerance
2.7$mm$), not the whole scene is interpret by planes. The tendency,
however, indicates when ample time permitted, the algorithms is capable
of explaining the scene in fine detail to portray a estimate of the
scene. More generally, as detection time increases, the scene area
that has not been fitted by planes decreases and the bandwidth increases.
The shade around the curves indicates the standard deviation of our
experimental data. The intersection of two solid curves (or dash curves)
is recommended to balance this trade-off: sufficient surface area
of the scene is described by planes under an optimal detection time
and bandwidth. In our results of dash curves, when $\sim$30$ms$
fitting time provided, $\sim$75\% surface area in the scene can be
represented while the bandwidth occupancy remains low ($\sim$10KB).
Fig.\ref{subfig:tradeoff2} shows the total number of planar fits
per image, which corresponds with the bandwidth. We also explore the
effect of the error tolerance to our SLAM system. In Fig \ref{subfig:tradeoff1},
the solid curves have error tolerance of 13.1$mm$ while the dash
curves 2.7$mm$. The actual average fitting error are 1.7$mm$ and
0.87$mm$ respectively. It is shown that for a same scene and a fixed
bandwidth capacity, the larger the error tolerance is, the less computational
resource is required. For example, at 10KB, the time comparison is
$\sim$60$ms$ versus $\sim$30$ms$. On the other hand, when computation
resources capacity is fixed, larger error tolerance allows more scene
fitted rate while only slightly larger bandwidth occupancy. These
results demonstrate the adaptive geometric sensitivity of our approach.
We manage to adjust the computation burden according to the bandwidth
capacity and fitting accuracy. Through our distributed graph-SLAM,
the agents can adaptively and accurately compress 3D surface data
and build maps.

A rate-distortion efficient map representation plays a significant
role in SLAM systems. To further explore the relationship between
fitting error and bandwidth, we develop a rate-distortion curve as
shown in Fig.\ref{fig:rate/distortion}. The distortion here is interpreted
as the fitting error, while the rate is dictated by the bandwidth
occupancy of the planar representation. We start with only fitting
blocks with the size of $40\times40$ pixels to images then continue
to reduce the block size to $3\times3$, while in this process, we
analyse the error and the bandwidth, as depicted in Fig.\ref{fig:rate/distortion}.
The \char`\"{}elbow\char`\"{} area (marked as red circle) of the curve
tells the optimal balance, with about 1.2$mm$ point-wise error and
around 40KB bandwidth requirement. It should be noted that, the computation
cost is not in the consideration here. Depending on the constraints
on the bandwidth capability, the reconstruction could be stopped at
any point, i.e. the algorithm allows scalability.

\subsection{Quadtree Level}

\begin{figure}[htbp]
\centering %
\begin{minipage}[t]{0.45\textwidth}%
\centering \includegraphics[width=3in,height=2.4in]{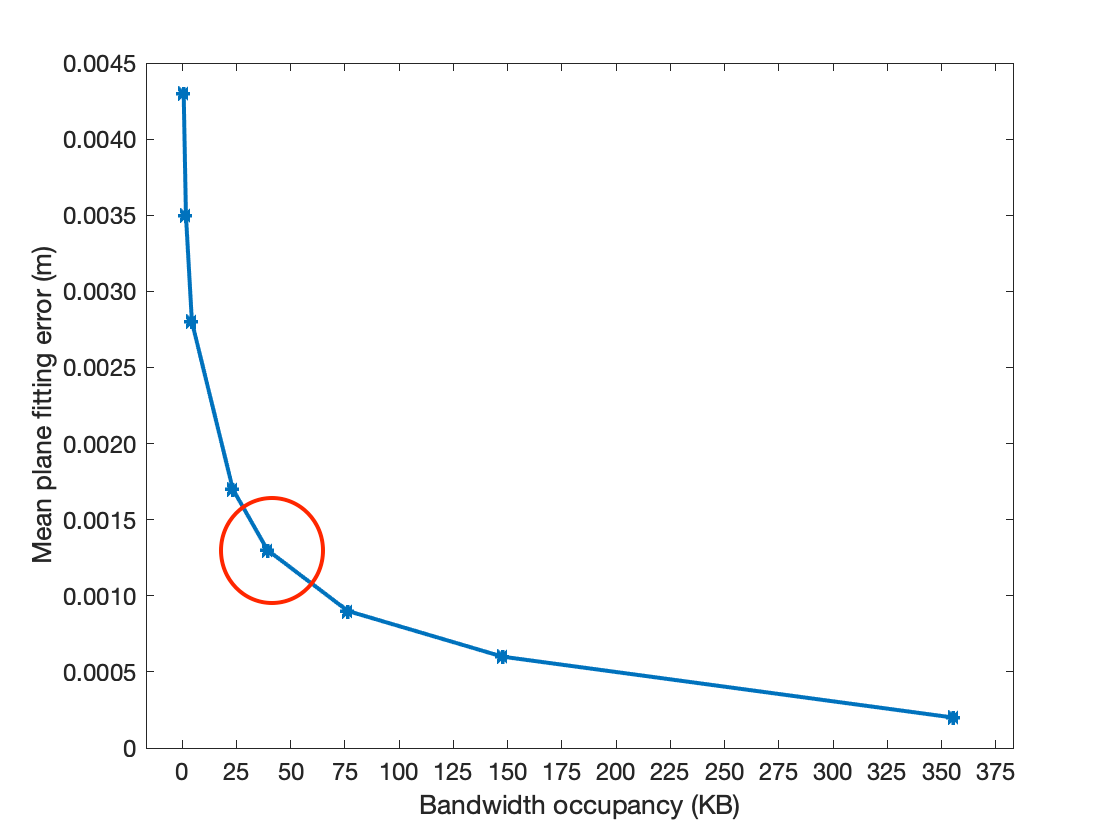}
\caption{Rate-distortion}
\label{fig:rate/distortion} %
\end{minipage}%
\begin{minipage}[t]{0.45\textwidth}%
\centering \includegraphics[width=3in,height=2.4in]{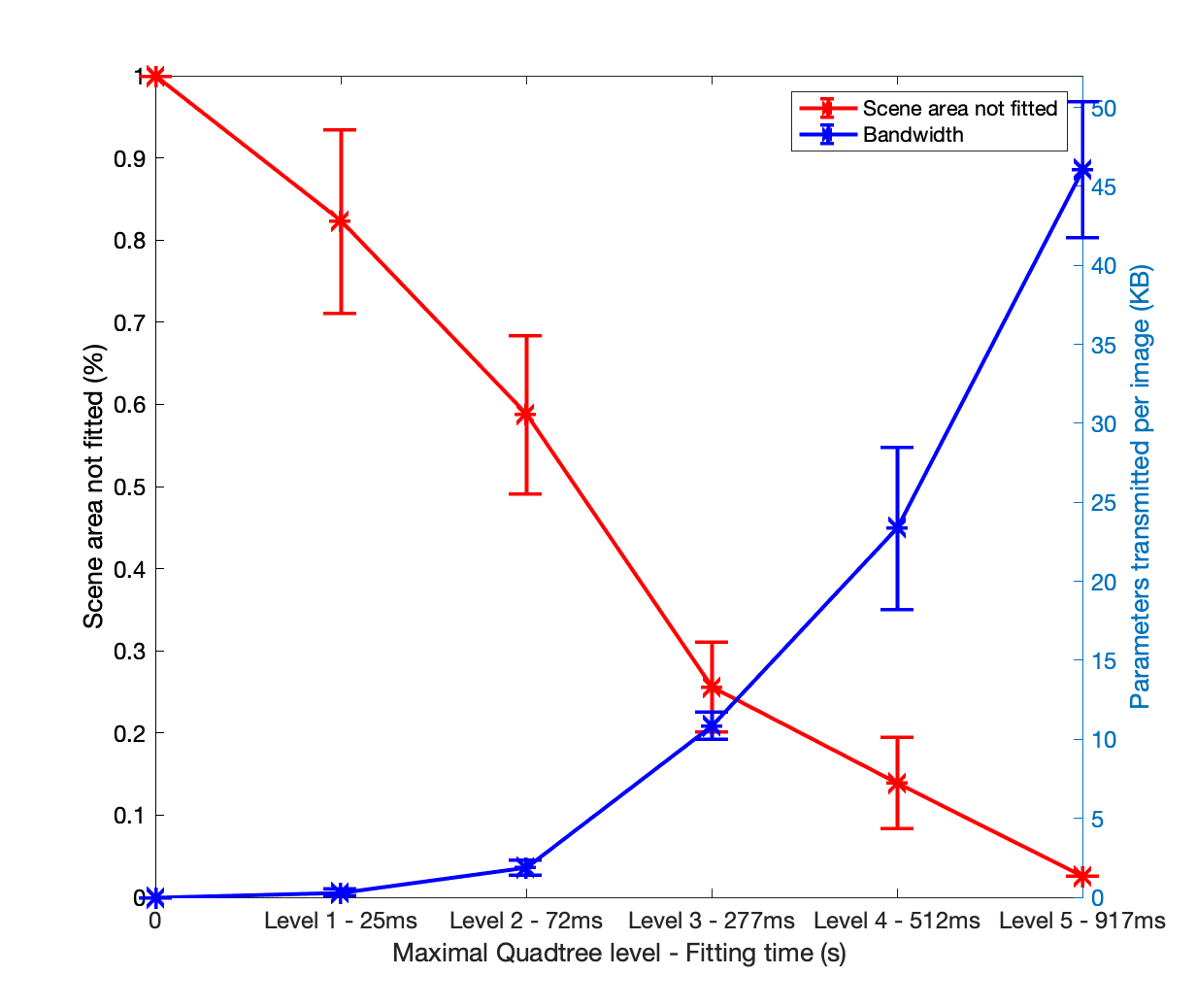}
\caption{Different detection Quadtree level}
\label{fig:Quadtree} %
\end{minipage}
\end{figure}

In Algorithm \ref{alg:fitting}, we explain that we enlarge the adaptability
of our approach by this fitting strategy: when there are tile candidates,
at different Quadtree levels our algorithms recursively divide the
planar tiles into available smaller cells and also recursively shrink
the current fitting size. We fit the awaiting tiles with with current
fitting blocks. Moreover, once a planar block fits a surface within
the error tolerance, we stop the dividing even if current block size
is larger than the minimal size. In this way, our approach promises
to reduce the computation and bandwidth occupancy by lowering the
Quadtree level. We choose 24$\times$24 pixels as our initial tile
size so the possible maximal Quadtree level is 5 (24-12-6-3-2). We
investigate 50 images with Quadtree level 5 and show the average experimental
statistics in Fig.\ref{fig:Quadtree}. Similar as before, the bars
show the standard deviation of data. Here, we restrict the fitting
Quadtree at different levels. Level 1 indicates all of the fitting
blocks are in the size of initial value (24$\times$24) and level
2 indicates only one recursion (for both tile candidates and fitting
size) is allowed. It is shown that as the Quadtree goes deeper, less
surface is left unfitted and meanwhile more bandwidth is consumed.
We also denote the surface fitting time for each level in the horizontal
axis. As we can see, for images revoking all recursion levels, the
scene fitted rate and the bandwidth occupancy are balanced when the
average Quadtree level is between 3 and 4. Also level 5 has the very
minor contribution to the scene fitted rate while still has a major
impact on the bandwidth. It is quite understandable considering that
level 5 contains the smallest fitting tiles but for each tile it still
takes 8 floats to define. Evidently, trimming the Quadtree to lower
levels can help save computational cost, as well as bandwidth load.
Note that here we show the results for images revoking all levels,
however, in the real-life scene fitting process it is not always the
case.

\subsection{Loop Closure}

Collaborative mapping in multi-agent environments requires each agent
to share information about their past and current states. To detect
loop closures in multi-robot contexts one robot needs to transfer its keyframe data
to other agents and also hold keyframes from other agents. It is challenging
to share raw sensor data like point cloud or other large amounts of
data among the agents in a distributed SLAM system. Hence, transferring
limited-size data is critical when dealing with a large collection
of robots. We reconstruct one of the hallways on the 2nd floor of
the EPIC building at the UNC Charlotte campus and show our comparison
in Table \ref{tab:loop closure data size}. In total 189 keyframes
are generated for this map of a 15$m\times$2.4$m$ hallway. It takes
221.48 MB of memory to represent this map if point cloud were used
but only 7.48 MB if plane fits were used, with nearly 30 times of
resource saving. The amount of data to be compared are 58,060,800
pixels and 245,249 plane fits. The reduction shown here is very significant
considering that in a distributed system, the resources consumption
has polynomial growth with respect to the number of agents.

\begin{table}[h]
\noindent \centering{}%
\begin{tabular}{lccc}
\hline 
 & depth image  & compact representation  & \tabularnewline
\hline 
\hline 
amount of data  & 58.06M pixels  & 0.25M plane fits  & \tabularnewline
\hline 
memory  & 221.48 MB  & 7.48 MB  & \tabularnewline
\hline 
\end{tabular}\vspace{1.5mm}
 \caption{Resources required to describe a hallway map of 189 keyframes}
\label{tab:loop closure data size} 
\end{table}

\begin{figure*}[htbp]
\centering \subfigure[]{ %
\begin{minipage}[t]{0.45\textwidth}%
\centering %
\fbox{\includegraphics[width=2.2in,height=2.2in]{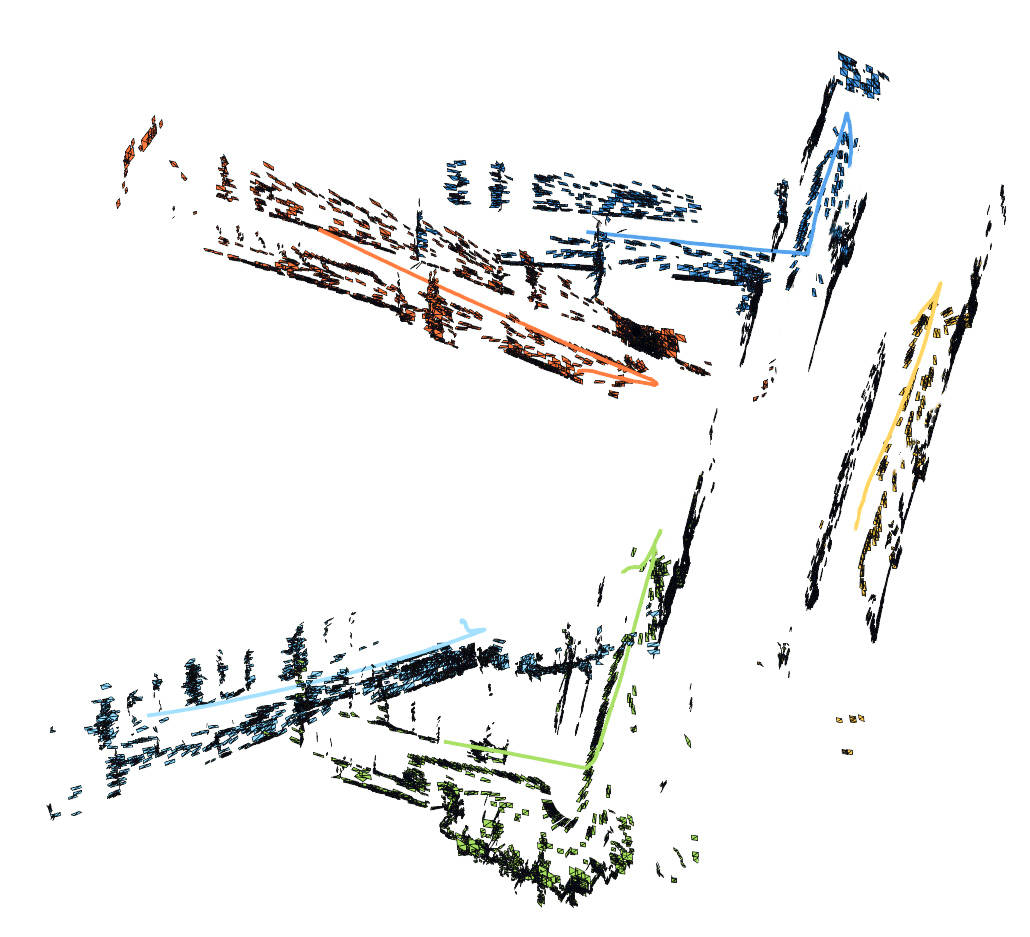}} %
\end{minipage}} \subfigure[]{ 
\begin{minipage}[t]{0.45\textwidth}%
\centering %
\fbox{\includegraphics[width=2.2in,height=2.2in]{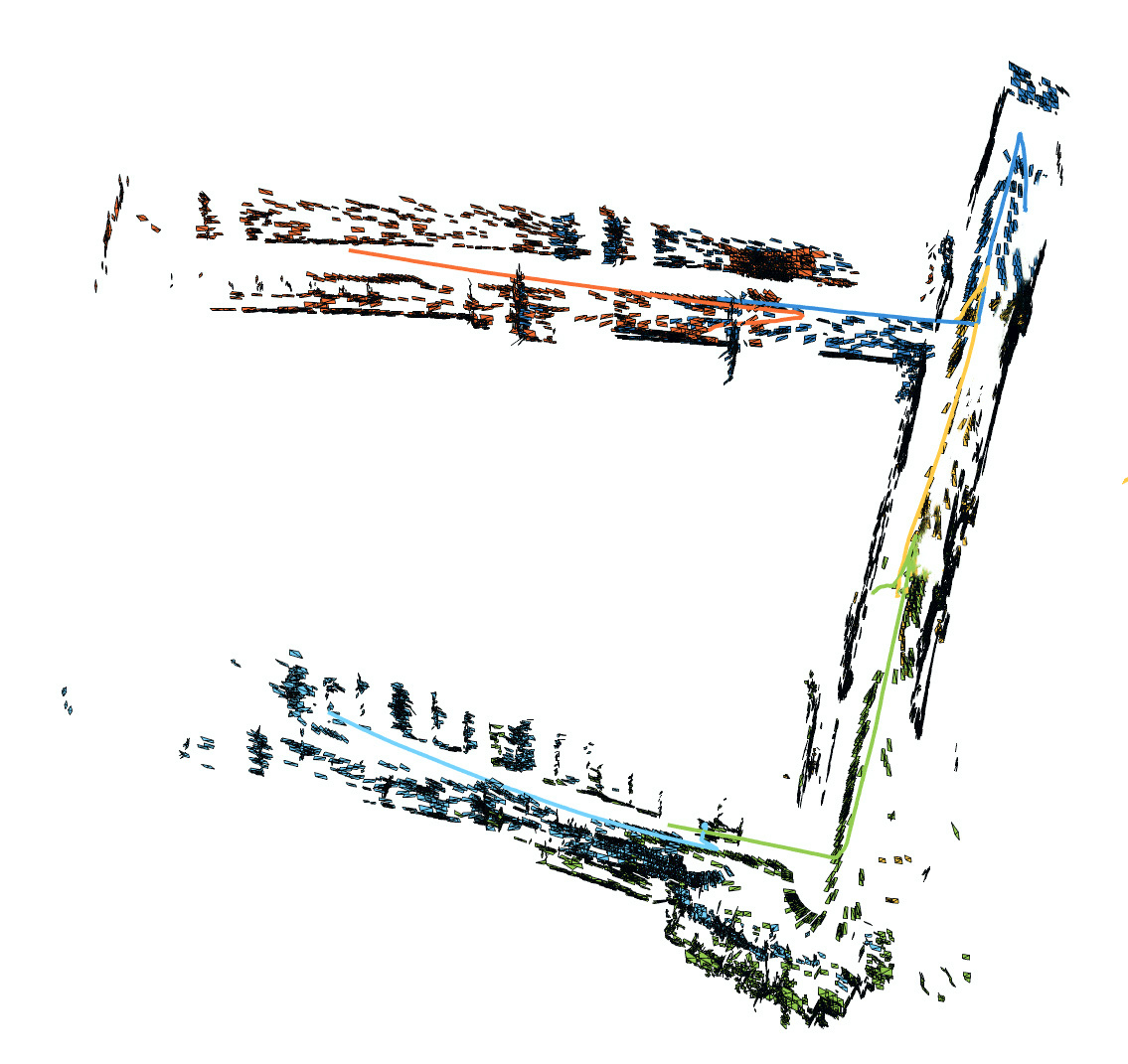}} %
\end{minipage}} \caption{The planar representation of dense 3D reconstructions without loop
closure (left) and with loop closure (right). Different maps with
visual trajectories are color-coded and down-sampled for better visualization.\label{fig:loopclosure}}
\end{figure*}

The map loop closures in our approach are generated by the identification
of overlapping scenes in shared plane sets. Fig.\ref{fig:loopclosure}
shows 4 loop closures among 5 maps generated by our robots when running
on the 2nd floor of the EPIC building. The hallway (including the
wall and the floor) is described in planar fits. All maps and corresponding
trajectories of agents are color-coded. Even a limited number of loop
closures or range measurements can provide strong ties between the
agents. If two agents, for example, the orange and the cyan in Fig.\ref{fig:loopclosure},
never communicate directly, it is quiet feasible for them to be connected
though a mutual neighbor (or chain of neighbors) like the green. In
benefit of this loop closure strategy, our approach has the potential
to construct observation with non-overlapping data from agents. Two
non-overlapping parts of a flat surface may share the same plane representation
coefficients (e.g., Hessian normal form), thus can be detected as
the same surface without observation of overlapping.

\section{Conclusion}


Using an RGBD sensor on-board turtlebots, we have demonstrated our approach
of adaptive compression for 3D surface data and map-building capability
as an extension to the DVO SLAM to create a new form of graph-SLAM. Free choice of the deployment of frontend and backend applications allows this new form of SLAM to take on realizations as either centralized or decentralized SLAM systems as required by the application. We have showed how the SLAM operations can be distributed across a frontend and backend system. The frontend processes the observations into
compact planar representations and then packages the information for
the backend system without requiring high-bandwidth communications.
The frontend maintains a short term history which notifies the backend
of potential loop-closures. The backend system then operates as the
workhorse in that it has additional computational resources available
to maintain a history of observations and confirm loop-closures. The
frontend system could be extended to multiple front-end system therefore
allowing multiple view points to be shared in a collaborative effort.
Our approach minimizes memory burden by supplying a compact data representation
which ensures computationally efficiency by adapting according to
processing availability.

Representing the observed scene with compact planar equations naturally
compresses the required information for SLAM and map-building. Although it takes very short computation time to fit surfaces with planes and introduces approximation uncertainty, by representing the 3D scene data with planar representations, we manage to minimize the computational burden of operating on dense
point clouds. With this approach, we're able to quickly integrate
a scene by operating on the planar equations to determine how the
planes interact with one another such as if they intersect to form
edges or corners as well as determining concave or convexity. Understanding
this provides insight on what potential objects could exist according
to the conditions of the observed planes, which also helps identify characteristic about the environment
such as open spaces, tight corridors, or obstructions.

Another notable contribution of this work is the adaptive compression capability.
SLAM algorithms have always sought for computational efficiency. Our
proposed method adapts according to the user's time and bandwidth
requirements. If ample time is permitted, the algorithms have the
opportunity to interrogate the scene in finer detail otherwise larger
generalizations are made that portray a rough estimate of the scene.
This allows the architecture to adapt rather than become overwhelmed
with sensory data. The algorithms also incorporate detected planar
regions as landmarks and use them to improve depth estimation, leading
to improvements in both camera tracking and dense reconstruction accuracy.
The estimated planar regions also provide invaluable semantic information
about the scene.

 \bibliographystyle{spiebib}
\bibliography{2020_SPIE_NetworkedGraphSLAM}

\end{document}